\pdfoutput=1

\documentclass[11pt]{article}

\usepackage[]{ACL2023}

\usepackage{times}
\usepackage{latexsym}

\usepackage{scalefnt}
\usepackage{amsfonts}
\usepackage{amsmath,bm}
\usepackage{mathrsfs}
\usepackage{multirow}
\usepackage{verbatim}
\usepackage{amssymb}
\usepackage{booktabs}
\usepackage{subcaption}
\usepackage{graphicx}
\usepackage{colortbl}

\usepackage[T1]{fontenc}

\usepackage[utf8]{inputenc}

\usepackage{microtype}

\usepackage{inconsolata}

\title{DecompEval: Evaluating Generated Texts as Unsupervised Decomposed Question Answering}

\author{Pei Ke$^1$, Fei Huang$^1$, Fei Mi$^2$, Yasheng Wang$^2$, Qun Liu$^2$, Xiaoyan Zhu$^1$, Minlie Huang$^1$\thanks{\quad Corresponding author} \\
\small $^1$The CoAI Group, DCST, Institute for Artificial Intelligence, State Key Lab of Intelligent Technology and Systems, \\
\small Beijing National Research Center for Information Science and Technology, Tsinghua University, Beijing 100084, China \\
\small $^2$Huawei Noah’s Ark Lab, China \\
\tt\small kepei1106@outlook.com, \ f-huang18@mails.tsinghua.edu.cn \\ 
\tt\small \{mifei2, wangyasheng, qun.liu\}@huawei.com, \{zxy-dcs, aihuang\}@tsinghua.edu.cn \\}

\begin{document}
\maketitle
\begin{abstract}

Existing evaluation metrics for natural language generation (NLG) tasks face the challenges on generalization ability and interpretability. Specifically, most of the well-performed metrics are required to train on evaluation datasets of specific NLG tasks and evaluation dimensions, which may cause over-fitting to task-specific datasets. Furthermore, existing metrics only provide an evaluation score for each dimension without revealing the evidence to interpret how this score is obtained. To deal with these challenges, we propose a simple yet effective metric called DecompEval. This metric formulates NLG evaluation as an instruction-style question answering task and utilizes instruction-tuned pre-trained language models (PLMs) without training on evaluation datasets, aiming to enhance the generalization ability. To make the evaluation process more interpretable, we decompose our devised instruction-style question about the quality of generated texts into the subquestions that measure the quality of each sentence. The subquestions with their answers generated by PLMs are then recomposed as evidence to obtain the evaluation result. Experimental results show that DecompEval achieves state-of-the-art performance in untrained metrics for evaluating text summarization and dialogue generation, which also exhibits strong dimension-level / task-level generalization ability and interpretability\footnote{The codes are available at \url{https://github.com/kepei1106/DecompEval}.}.



\end{abstract}

\section{Introduction}


Recently, pre-trained language models (PLMs) such as GPT \cite{brown2020gpt3}, BART \cite{lewis2020bart}, and T5 \cite{raffel2020t5} have achieved promising performance in natural language generation (NLG) tasks, such as text summarization \cite{zhang2020pegasus} and dialogue generation \cite{zhang2020dialogpt}. As the quality of generated texts gradually approaches that of human-written texts, there is an increasing demand for automatic evaluation metrics of generated texts.


However, existing evaluation metrics are still struggling to measure the quality of generated texts accurately. Traditional metrics such as BLEU \cite{papi2002bleu}, METEOR \cite{banerjee2005meteor}, and ROUGE \cite{lin2004rouge} rely on n-gram overlap between generated texts and reference texts, which fail to detect the issues in the content of generated texts \cite{gehrmann2022repairing}. 
Recent works resort to model-based evaluation metrics to compute the similarity between generated texts and reference texts based on contextual representations from pre-trained models \cite{zhao2019moverscore,zhang2020bertscore} or adopt the score of language modeling \cite{yuan2021bartscore} / masked language modeling \cite{ke2022ctrleval,colombo2022infolm} for evaluation.
Other works choose to train evaluation models on the evaluation datasets to fit human scores \cite{shen2017style,sellam2020bleurt} or distinguish human-written texts from negative samples \cite{guan2020union,zhong2022unieval}, aiming to obtain higher correlations with human judgments in various evaluation dimensions (such as coherence and consistency) of specific datasets.

We argue that there are two main challenges in building an evaluation metric for text generation: 1) \textbf{Generalization Ability}: Most of the existing metrics that have high correlations with human judgments on evaluation datasets are directly trained on the corresponding datasets 
\cite{sellam2020bleurt,guan2020union,zhong2022unieval}. This may result in over-fitting to task-specific data and harm their generalization ability to other NLG tasks and dimensions \cite{ke2022ctrleval}.
2) \textbf{Interpretability}: Although recently proposed evaluation metrics can measure the quality of generated texts from multiple dimensions, they only provide an evaluation score for each dimension without giving evidence to interpret how they predict this score \cite{ke2022ctrleval,zhong2022unieval}.

To deal with these challenges, we propose a simple yet effective evaluation metric called DecompEval. \textbf{Firstly}, to improve the generalization ability, we formulate NLG evaluation as an instruction-style question answering (QA) task,
and utilize instruction-tuned pre-trained language models \cite{chung2022flant5} to solve this task without training on task-specific data. The instruction-style question consists of an instruction, the input of NLG evaluation, and a yes/no question, e.g., "\textit{Answer the following yes/no question ... Is this a coherent response given the dialogue history?}" for the evaluation of coherence in dialogue generation, where the specific evaluation input is omitted.
\textbf{Secondly}, we propose a question decomposition strategy to make the evaluation process more interpretable, instead of directly making instruction-tuned PLMs answer the original question. This strategy decomposes the question 
into the subquestions which sequentially evaluate the corresponding dimension of each sentence in the generated texts. 
Then, we recompose these subquestions with their answers generated by the PLM as evidence to make the PLM answer the original question, which is used to compute the final evaluation result.
The evidence can promote the understanding of the evaluation process by indicating the potential problematic sentences that affect the evaluation score.

Our main contributions are as follows:

\begin{itemize}
    \item We propose an evaluation metric called DecompEval, which formulates NLG evaluation as an instruction-style QA task, and solves it with instruction-tuned PLMs via question decomposition.
    \item We conduct experiments on the benchmark datasets for evaluating text summarization and dialogue generation. Experimental results show that DecompEval can achieve state-of-the-art performance in untrained metrics. 
    
    \item We empirically show that DecompEval can generalize to other evaluation dimensions and tasks (such as data-to-text generation) better than all the baselines, while improving the interpretability via decomposed subquestions with their answers.
\end{itemize}

\begin{figure*}[!t]
  \centering
  \includegraphics[width=1.0\linewidth]{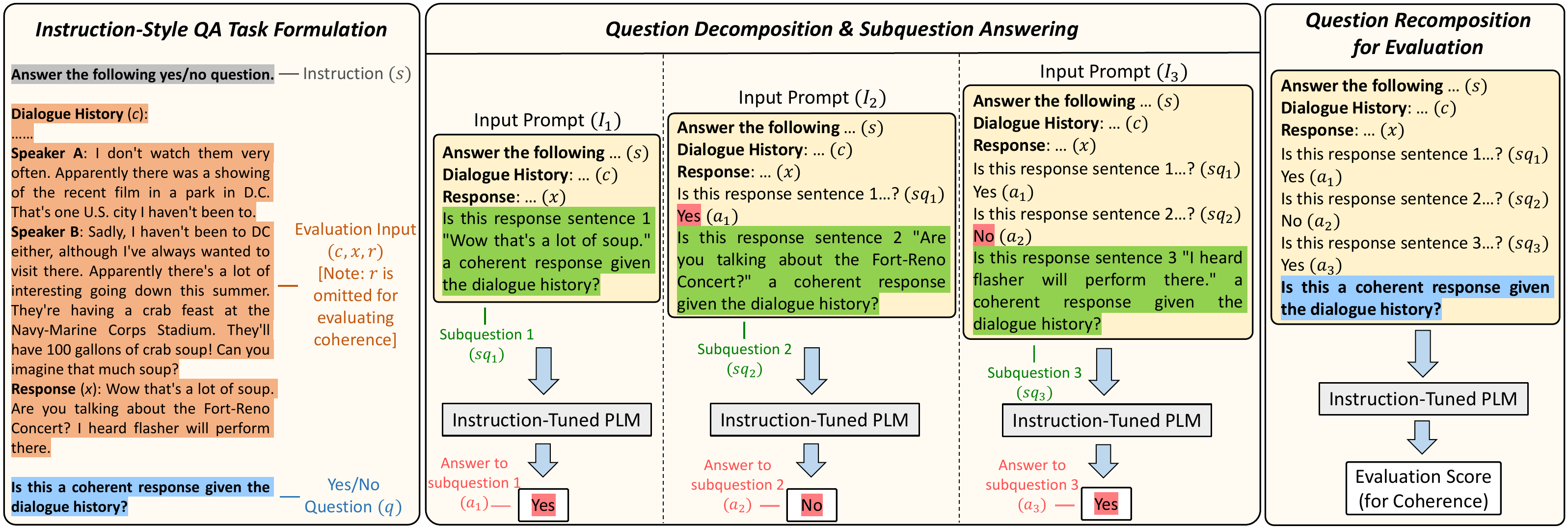}
  \caption{The overview of DecompEval. We take the evaluation of coherence in dialogue generation as an example. \textbf{Left}: The input of evaluation is formulated as an instruction-style question, which contains an instruction, a tuple of evaluation inputs, and a yes/no question about the quality of generated responses. \textbf{Medium}: The instruction-style question is decomposed into subquestions according to sentences. At each step, the instruction-tuned PLM generates an answer to the current subquestion based on the input prompt. Then, the answer becomes the constituent of the input prompt at the next step. \textbf{Right}: The instruction-tuned PLM recomposes all the subquestions with their answers to answer the original question and acquire the evaluation result.}
  \label{fig:overview}
\end{figure*}

\section{Related Work}

\subsection{Evaluation for Language Generation}

Evaluation is a long-standing task in the field of NLG \cite{celi2020evalsurvey}, which becomes more critical with the rapid development of PLMs. There are two main categories of automatic evaluation metrics, i.e., untrained and trained metrics \cite{sai2020ddplus}. Untrained metrics without training on specific datasets of evaluation tasks or related tasks aim to measure the relationship among source texts, generated texts, and reference texts via n-gram overlap \cite{papi2002bleu,banerjee2005meteor,lin2004rouge}, semantic similarity \cite{zhao2019moverscore,zhang2020bertscore}, or language modeling / masked language modeling scores \cite{yuan2021bartscore,ke2022ctrleval,colombo2022infolm}. In comparison, trained metrics are commonly trained on the evaluation datasets to fit human scores \cite{shen2017style,sellam2020bleurt} or distinguish human-written texts from negative samples \cite{guan2020union,zhong2022unieval}, aiming to achieve higher correlations with human judgments on specific datasets. Among these metrics, there are some similar works which re-frame NLG evaluation as QA tasks and adopt the generated answers or generation probabilities as evaluation results \cite{daniel2021qaeval,zhong2022unieval}.

The most similar work to our method is UniEval \cite{zhong2022unieval}. UniEval re-frames NLG evaluation as a Boolean QA task and trains the evaluation model on the pseudo data constructed from the evaluation dataset and other related datasets in a unified Boolean QA format. Compared with UniEval, our method is untrained since we 
transform NLG evaluation to an instruction-style QA task that can be solved by instruction-tuned PLMs without further training. Also, our method can provide some evidence (i.e., the answers to decomposed subquestions) to interpret how the model reaches the evaluation result, instead of only providing a final evaluation score.


\subsection{Instruction-Tuned Pre-Trained Models}

Instruction learning \cite{weller2020instructlearning} which trains PLMs to follow human instructions has attracted much attention recently since it shows the strong zero-shot cross-task generalization ability. To improve instruction understanding, existing works adopt instruction tuning \cite{wei2022flan} which trains PLMs on massive tasks described via instructions with multi-task learning, such as FLAN \cite{wei2022flan,chung2022flant5}, T0 \cite{sanh2022t0}, and InstructGPT \cite{ouyang2022instructgpt}. Other works systematically study instruction tuning in specific areas such as dialogue systems \cite{gupta2022instructdial} and multi-modal learning \cite{xu2022multiinstruct}.

In comparison, our work is the first to explore the potential of instruction-tuned PLMs in the evaluation of NLG without further training. We show that equipped with well-designed input prompts and suitable question decomposition, instruction-tuned PLMs can sequentially measure the quality of each sentence and finally recompose all the subquestions with their answers to obtain surprisingly great evaluation results in an unsupervised fashion.

\section{Method}

\subsection{Task Definition and Model Overview}
\label{sec:overview}

Given the context $c$, the model-generated text $x$, and the reference text $r$, our goal is to acquire the evaluation results from different individual dimensions, respectively. The context contains different contents in various NLG tasks. Also, the context and the reference may be omitted, which depend on the evaluation task and dimension. We assume that the generated text consists of $n$ sentences, i.e., $x=(x_1,x_2,\cdots,x_n)$.

As shown in Figure \ref{fig:overview}, our main idea is to formulate NLG evaluation as an instruction-style QA task and solve this task with instruction-tuned PLMs via question decomposition. Our proposed method consists of three steps. First of all, we transform the input of NLG evaluation into an instruction-style question which contains an instruction $s$, the input of evaluation tasks $(c,x,r)$, and a yes/no question $q$ for each dimension (\S\ref{sec:mhqaformulation}). Then, we decompose this question into the subquestions $\{sq_t\}_{t=1}^n$, which evaluate each sentence $x_t(1\leq t\leq n)$ in the generated text $x$ respectively and acquire the answers $\{a_t\}_{t=1}^n$ to these subquestions via the instruction-tuned PLM $P_\theta$ (\S\ref{sec:questiondecomp}). The answer to each subquestion is appended to the input prompt of the PLM, which may help to solve subsequent subquestions as in-context examples. Finally, we recompose all the subquestions with their answers as evidence and make the instruction-tuned PLM answer the original question, which can be used to compute the evaluation result (\S\ref{sec:questionrecomp}).


\subsection{Instruction-Style QA Task Formulation}
\label{sec:mhqaformulation}

To improve the generalization ability of evaluation metrics, we formulate NLG evaluation as an instruction-style QA task that can be solved by instruction-tuned PLMs in an unsupervised fashion. As shown in Figure \ref{fig:overview}, the instruction-style question contains three parts:
\begin{itemize}
    \item \textbf{Instruction}: The design of instructions depends on the data format of instruction-tuned PLMs. In this paper, we adopt yes/no questions \cite{zhong2022unieval} to measure the quality of generated texts. Thus, we follow \citet{chung2022flant5} to devise the instruction as $s=$"\textit{Answer the following yes/no question.}".
    \item \textbf{Evaluation Input}: The original input $(c,x,r)$ for NLG evaluation mentioned in \S\ref{sec:overview} are incorporated with task-specific descriptive texts. For example, we add the text "\textit{dialogue history:}", "\textit{response:}", and "\textit{reference:}" before $c$, $x$, and $r$ respectively for evaluating dialogue generation.
    \item \textbf{Yes/No Question}: We finally devise a yes/no question to assess the specific dimension of generated texts. For example, the yes/no question assessing the coherence of generated texts in dialogue generation is $q=$"\textit{Is this a coherent response given the dialogue history?}".
\end{itemize}



\subsection{Question Decomposition and Subquestion Answering}
\label{sec:questiondecomp}

To interpret how the model predicts the evaluation score, we devise a question decomposition strategy inspired by the existing works in the QA community \cite{min2019decomprc,perez2020decomp,zhou2022l2mprompt},
rather than force the instruction-tuned PLM to answer the original question directly. 
This strategy splits the generated text based on sentences and sequentially queries the quality of each sentence via subquestions. The subquestions with their answers generated by the PLM are expected to act as evidence to illustrate how the PLM arrives at the final evaluation score. We simply select sentences as the decomposition criterion instead of using external off-the-shelf models \cite{perez2020decomp,daniel2021qaeval} because sentences are shown to be important basic units for deriving the evaluation result of the whole generated text \cite{amplayo2022smart}.

Specifically, to answer the subquestion $sq_t (1\leq t\leq n)$ for measuring the quality of the $t$-th sentence $x_t$, we combine the instruction $s$, the evaluation input $(c,x,r)$, the previous subquestions with their answers $\{(sq_j,a_j)\}_{j=1}^{t-1}$, and the current subquestion $sq_t$ as the input prompt $I_t=\left(s,c,x,r,\{(sq_j,a_j)\}_{j=1}^{t-1},sq_t\right)$. Then, we compare the generation probability of "\textit{yes}" / "\textit{no}" from the instruction-tuned PLM to determine the answer:
\begin{eqnarray}
a_t & = &\left\{
\begin{aligned}
\textrm{yes} , \quad P_\theta(\textrm{yes}|I_t)> P_\theta(\textrm{no}|I_t) \\
\textrm{no} , \quad P_\theta(\textrm{yes}|I_t)\leq P_\theta(\textrm{no}|I_t)
\end{aligned}
\right. \\
t & = & 1,2,\cdots,n \notag
\end{eqnarray}

The answer $a_t$ is appended to the current input prompt $I_t$, which becomes the in-context examples of $I_{t+1}$ helping to solve the next subquestion $sq_{t+1}$. All these subquestions with their answers can serve as evidence to
improve the interpretability by indicating potential low-quality sentences in the generated text that affect the evaluation score.

\subsection{Question Recomposition for Evaluation}
\label{sec:questionrecomp}

To recompose all the subquestions with their answers to acquire the final evaluation result, we append the original yes/no question mentioned in \S\ref{sec:mhqaformulation} to the end of the last subquestion and its answer. The instruction-tuned PLM is expected to leverage all these information as evidence to answer the original question and obtain the evaluation result.

Specifically, given the instruction $s$, the evaluation input $(c,x,r)$, all the subquestions with their answers $\{(sq_t,a_t)\}_{t=1}^n$, and the original question $q$ as the input prompt, we compute the evaluation score using the generation probability of answer words (i.e., yes and no) from the instruction-tuned PLM \cite{ke2022ctrleval,zhong2022unieval}:
\begin{align}
    f(l) & =  P_\theta(l|s,c,x,r,\{(sq_t,a_t)\}_{t=1}^n,q) \\
    score & = \frac{f(l=\textrm{yes})}{f(l=\textrm{yes})+f(l=\textrm{no})}
\end{align}

\section{Experiment}

\subsection{Dataset}

We follow \citet{zhong2022unieval} to adopt two benchmark datasets to test the performance of DecompEval. The statistics of these datasets are shown in Table \ref{tab:datastat}. 

\noindent \textbf{SummEval} \cite{fabbri2021summeval}: This dataset is a benchmark for evaluation metrics of text summarization. It covers the generated summaries from recent summarization models on the CNN/DailyMail (CNNDM) dataset \cite{hermann2015cnndm}. For each generated summary, it provides the human scores from four dimensions including fluency, coherence, consistency, and relevance.

\noindent \textbf{Topical-Chat} \cite{gop2019topicalchat}: This dataset is a benchmark for knowledge-grounded dialogue generation. \citet{mehri2020usr} collects human annotations for the models trained on Topical-Chat. For each generated response, it provides the human scores from five dimensions\footnote{We use the description of dimensions in the existing work \cite{zhong2022unieval} for fair comparison, which is slightly different from the original paper \cite{mehri2020usr}.} including naturalness, coherence, engagingness, groundedness, and understandability. Following \citet{zhong2022unieval}, we use the first four dimensions in the main result (\S\ref{sec:mainresult}) and the last dimension to test the generalization ability (\S\ref{sec:generalization}).



\begin{table} [!t]
\centering
\scriptsize
\setlength{\tabcolsep}{1.3mm}{
\begin{tabular}{l|c|c|c|c}
\toprule
Dataset & Task & \#Samples & \#Dimensions & Length \\
\midrule
SummEval & Text Summarization & 1,600 & 4 & 63.7    \\
Topical-Chat & Dialogue Generation & 360 & 5 & 22.9  \\
\bottomrule
\end{tabular}}
\caption{Statistics of the benchmark datasets, including the task, the number of samples / dimensions, and the average length of generated texts.}
\label{tab:datastat}
\end{table}

\subsection{Implementation Detail}

We choose FLAN-T5 \cite{chung2022flant5} as our base model, which is obtained by training T5 \cite{raffel2020t5} on 1.8K tasks described via instructions\footnote{Although the instruction-tuning datasets of FLAN-T5 cover the CNNDM dataset \cite{chung2022flant5}, they do not include the generated summaries with human evaluation scores, ensuring no data leak in the experiment.}. We use FLAN-T5-XL with 3B parameters in the main result
and also explore other model scales in \S\ref{sec:modelscale}. We follow \citet{zhong2022unieval} to set the input length to be 1,024. We design the input prompts based on the data formats of FLAN-T5, the evaluation tasks and dimensions.
More details about the specific design of input prompts for each dataset / dimension and the sensitivity analysis are included in Appendix \ref{app:instruct}.


As for the evaluation on two datasets, we directly compute summary-level / turn-level evaluation scores for SummEval / Topical-Chat based on our method in most of the dimensions, respectively, except fluency / consistency on SummEval and engagingness on Topical-Chat. For these dimensions, we follow \citet{zhong2022unieval} to obtain the evaluation scores via averaging (for fluency / consistency on SummEval) \cite{laban2022summac} or cumulating (for engagingness on Topical-Chat) \cite{deng2021unified} individual evaluation results of constituent sentences for fair comparison.


\begin{table*} [!h]
\centering
\small
\begin{tabular}{l|c|c|c|c|c|c|c|c}
\toprule
Dimension  & \multicolumn{2}{c|}{Coherence} & \multicolumn{2}{c|}{Consistency} & \multicolumn{2}{c|}{Fluency} & \multicolumn{2}{c}{Relevance}  \\
\cmidrule{1-9}
Metric & $\rho$ & $\tau$ & $\rho$ & $\tau$ & $\rho$ & $\tau$ & $\rho$ & $\tau$ \\
\midrule
\multicolumn{9}{c}{\textit{Trained Metric (w/ Training on Data of Evaluation Tasks or Related Tasks)}} \\
\midrule
BARTScore (CNNDM)  & 0.448 & 0.342 & 0.382 & 0.315  & 0.356   & 0.292   & 0.356  & 0.273    \\
UniEval (Summ)  & \underline{0.575}  & \underline{0.442} &  0.446 &   0.371  & \underline{0.449}   & \underline{0.371}   & \underline{0.426}  &  \underline{0.325}  \\
\midrule
\midrule
\multicolumn{9}{c}{\textit{Untrained Metric (w/o Training on Data of Evaluation Tasks or Related Tasks)}} \\
\midrule
ROUGE-1  & 0.167 & 0.126 & 0.160 & 0.130  & 0.115 & 0.094  & 0.326 & 0.252    \\
ROUGE-2  & 0.184 & 0.139 & 0.187 & 0.155  & 0.159 & 0.128  & 0.290  & 0.219    \\
ROUGE-L  & 0.128 & 0.099 &  0.115 & 0.092 & 0.105 & 0.084  & 0.311 &  0.237   \\
MoverScore  & 0.159 & 0.118 & 0.157 & 0.127  & 0.129 & 0.105  & 0.318 & 0.244    \\
BERTScore  & 0.284 & 0.211 & 0.110 & 0.090  & 0.193  & 0.158  & 0.312 & 0.243    \\
BARTScore &  0.322 &  0.250  &  0.311  &  0.256   &  0.248  &  0.203  &  0.264  &  0.197   \\
CTRLEval &  0.217  &  0.164   &  0.301   &   0.247   & 0.132  &  0.107   &  0.196   &  0.152    \\
DecompEval (Ours) & \textbf{0.341}  & \textbf{0.256}  &  \underline{\textbf{0.455}} &  \underline{\textbf{0.378}}  &  \textbf{0.285}  & \textbf{0.233}  &  \textbf{0.355} &  \textbf{0.276}   \\
\bottomrule
\end{tabular}
\caption{Summary-level Spearman ($\rho$) and Kendall ($\tau$) correlations of coherence, consistency, fluency, and relevance on the SummEval dataset. The highest correlation
for each dimension achieved by untrained metrics is \textbf{bold}, while the highest correlation overall is \underline{underlined}.}
\label{tab:mainsum}
\end{table*}

\subsection{Baseline}

We choose several state-of-the-art untrained and trained metrics as our baselines:

\noindent \textbf{MoverScore} \cite{zhao2019moverscore}: This metric relies on Earth Mover’s Distance \cite{rubner2000emd} between generated texts and reference texts based on the contextual representations from PLMs.

\noindent \textbf{BERTScore} \cite{zhang2020bertscore}: This metric computes the similarity between generated texts and reference texts based the contextual representations from BERT \cite{devlin2019bert}.

\noindent \textbf{USR} \cite{mehri2020usr}: This metric combines the evaluation results of masked language models and dialogue retrieval models which are trained on the dialogue evaluation dataset.

\noindent \textbf{BARTScore} \cite{yuan2021bartscore}: This metric utilizes the generation probabilities of BART \cite{lewis2020bart} to measure the relationship among source texts, generated texts, and reference texts with different inputs and outputs. We use two variants \textbf{BARTScore} and \textbf{BARTScore (CNNDM)} in the original paper. The latter adopts BART fine-tuned on the CNNDM dataset as the base model.


\noindent \textbf{CTRLEval} \cite{ke2022ctrleval}: This metric formulates evaluation dimensions as multiple text infilling tasks and uses the ensemble of generation probabilities from PEGASUS \cite{zhang2020pegasus} as the evaluation results.


\noindent \textbf{UniEval} \cite{zhong2022unieval}: This metric re-frames NLG evaluation as a Boolean QA task. It conducts multi-task learning on the related datasets and continual learning on the dimensions of the evaluation dataset with a unified QA format. We use two variants \textbf{UniEval (Summ)} and \textbf{UniEval (Dial)} in the original paper, which are trained on all the dimensions of SummEval and the first four dimensions of Topical-Chat, respectively.

In addition, we also select traditional evaluation metrics based on n-gram overlap like BLEU \cite{papi2002bleu}, METEOR \cite{banerjee2005meteor}, and ROUGE \cite{lin2004rouge} as baselines. We directly re-print the experimental results of baselines if their original papers adopt the same benchmark datasets as ours. Otherwise, we implement the baselines based on the codes and model parameters released by the original papers.

\begin{table*} [!h]
\centering
\small
\begin{tabular}{l|c|c|c|c|c|c|c|c}
\toprule
Dimension  & \multicolumn{2}{c|}{Naturalness} & \multicolumn{2}{c|}{Coherence} & \multicolumn{2}{c|}{Engagingness} & \multicolumn{2}{c}{Groundedness}  \\
\cmidrule{1-9}
Metric & $r$ & $\rho$  & $r$ & $\rho$ & $r$ & $\rho$  & $r$ & $\rho$ \\
\midrule
\multicolumn{9}{c}{\textit{Trained Metric (w/ Training on Data of Evaluation Tasks or Related Tasks)}} \\
\midrule
USR  & 0.337 & 0.325 & 0.416 & 0.377 & 0.456 &  0.465 & 0.222 & 0.447    \\
UniEval (Dial)  & \underline{0.444} & \underline{0.514} & \underline{0.595} & \underline{0.613} & \underline{0.557} & \underline{0.605} & 0.536 & 0.575  \\
\midrule
\midrule
\multicolumn{9}{c}{\textit{Untrained Metric (w/o Training on Data of Evaluation Tasks or Related Tasks)}} \\
\midrule
BLEU-1  & 0.161 & 0.133 & 0.210 & 0.223 & 0.314 & 0.334 & 0.289 & 0.303  \\
BLEU-4  & 0.180 & 0.175 & 0.131 & 0.235 & 0.232 & 0.316 & 0.213 & 0.310    \\
ROUGE-L  & 0.176 & 0.146 & 0.193 & 0.203 & 0.295 & 0.300 & 0.310 & 0.327   \\
METEOR  & 0.212 & 0.191 & 0.250 & 0.302 & 0.367 & 0.439 & 0.333 & 0.391   \\
MoverScore &  0.169 & 0.170 & 0.247 & 0.259 & 0.275 & 0.269 & 0.198 & 0.147 \\
BERTScore  & 0.226 & 0.209 & 0.214 & 0.233 & 0.317 & 0.335 & 0.291 
& 0.317    \\
BARTScore & 0.287  & 0.266  & 0.251  & 0.225  &  0.411  & 0.406  & 0.226  & 0.205  \\
CTRLEval & 0.303  &  0.254  &  0.337  &  0.313 &  0.422  & 0.412   &  0.242  &  0.251   \\
DecompEval (Ours) &  \textbf{0.410}  & \textbf{0.435}  & \textbf{0.434} & \textbf{0.435}  &  \textbf{0.453}  & \textbf{0.467}  & \underline{\textbf{0.646}} & \underline{\textbf{0.659}}    \\
\bottomrule
\end{tabular}
\caption{Turn-level Pearson ($r$) and Spearman ($\rho$) correlations of naturalness, coherence, engagingness, and groundedness on the Topical-Chat dataset. The highest correlation
for each dimension achieved by untrained metrics is \textbf{bold}, while the highest correlation overall is \underline{underlined}.}
\label{tab:maindial}
\end{table*}

\subsection{Main Result}
\label{sec:mainresult}

Following \citet{liu2021explainboard} and \citet{zhong2022unieval}, we adopt summary-level Spearman ($\rho$) and Kendall ($\tau$) correlation coefficients between human judgments and automatic metrics to assess the performance on the SummEval dataset. The results in Table \ref{tab:mainsum} show that DecompEval achieves state-of-the-art performance in untrained metrics, indicating the effectiveness of our proposed instruction-style QA formulation and question decomposition method. 
Especially, DecompEval can even beat the best-performing trained metric UniEval (Summ) in the dimension of consistency, which shows the potential of instruction-tuned PLMs in the evaluation of generated texts.

We also conduct experiments on the Topical-Chat dataset and report turn-level Pearson ($r$) / Spearman ($\rho$) correlation coefficients in Table \ref{tab:maindial} as the existing works \cite{mehri2020usr,zhong2022unieval} do. Similarly, DecompEval beats all the untrained baselines and even outperforms the trained baseline USR in most of the dimensions. This indicates that DecompEval can successfully adapt to the evaluation of dialogue generation without training on specific datasets.
We also find that DecompEval can outperform UniEval (Dial) in the dimension of groundedness. We conjecture that DecompEval may be good at measuring the consistency between generated texts and contexts, thereby performing extremely well on consistency in text summarization and groundedness in dialogue generation.

\subsection{Generalization Ability}
\label{sec:generalization}

Generalization ability is essential because new evaluation dimensions and tasks may emerge without sufficient data. Thus, we study whether DecompEval can generalize at the dimension / task level better than untrained and trained baselines.


\subsubsection{Generalization to Other Dimensions}

\begin{figure}[!t]
  \centering
  \includegraphics[width=1.0\linewidth]{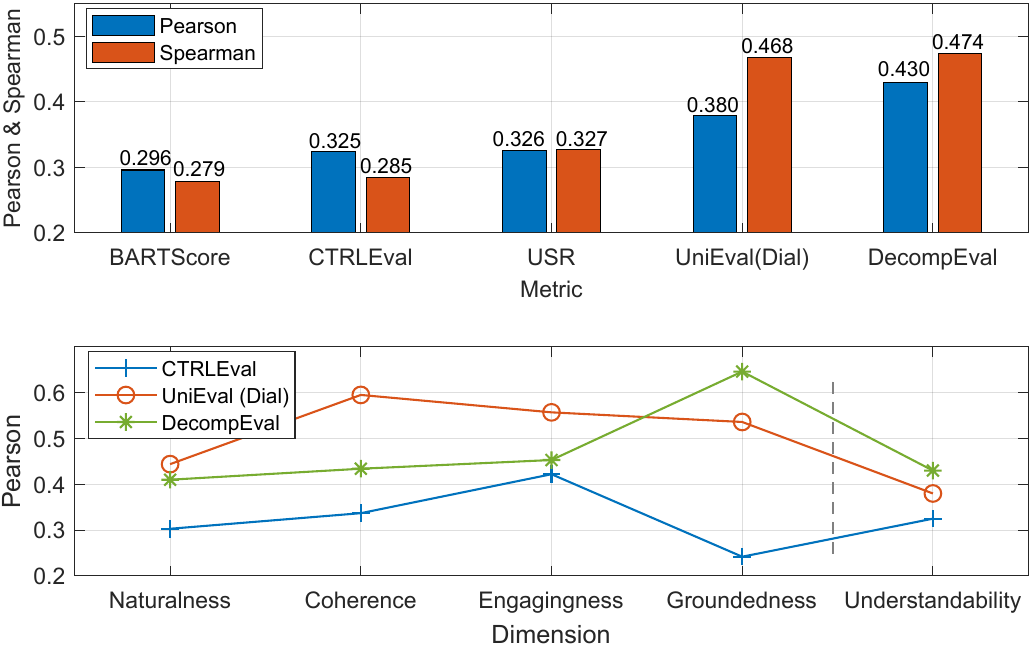}
  \caption{\textbf{Top}: Pearson and Spearman correlations of different metrics in the dimention of understandability. \textbf{Bottom}: Pearson correlation of CTRLEval, UniEval (Dial), and DecompEval in all the five dimensions of Topical-Chat.}
  \label{fig:genelizationdimension}
\end{figure}

To compare the performance of DecompEval and untrained / trained baselines on other dimensions,
we follow \citet{zhong2022unieval} to adopt the dimension of understandability on the Topical-Chat dataset to conduct experiments. 

The results in the top of Figure \ref{fig:genelizationdimension} show that DecompEval can outperform all the competitive untrained / trained baselines and achieve best performance in the dimension of understandability, which shows its strong dimension-level generalization ability. From the bottom of Figure \ref{fig:genelizationdimension}, we can observe that DecompEval maintains stable performance in all these dimensions.
In comparison, the trained baseline UniEval (Dial) which is trained on the first four dimensions of Topical-Chat except understandability cannot surpass DecompEval in the evaluation of understandability. 
The performance of UniEval (Dial) also degrades obviously in understandability compared with the other dimensions,
which demonstrates the potential side-effect of over-fitting to specific dimensions.



\subsubsection{Generalization to Other NLG tasks}

To investigate how DecompEval performs compared with untrained / trained baselines in other NLG tasks in addition to text summarization and dialogue generation,
we follow \citet{yuan2021bartscore} and \citet{zhong2022unieval} to adopt two data-to-text generation datasets SFRES and SFHOT \cite{wen2015d2teval}. These two datasets cover generated texts from structured data in the domain of restaurants and hotels. For each generated text, they provide human scores from two dimensions, i.e., naturalness and informativeness. The number of samples in SFRES / SFHOT is 1,181 / 875, respectively.

The results are shown in Table \ref{tab:generalizationd2t}. Our proposed metric DecompEval can still achieve state-of-the-art performance in untrained metrics and outperform the trained baselines in most of the dimensions. Thus, we believe that DecompEval can successfully improve the generalization ability to multiple NLG tasks via the full utilization of the instruction-tuned PLM without further training. We also illustrate the average of Spearman correlation coefficients in all the dimensions of each dataset in Table \ref{tab:generalizationtask}. Compared with our proposed metric, UniEval (Summ) and UniEval (Dial), as the best-performing trained metrics on the SummEval and Topical-Chat datasets, respectively, obtain obviously worse performance on the evaluation datasets which they are not trained on, indicating limited task-level generalization ability.



\begin{table} [!t]
\centering
\small
\setlength{\tabcolsep}{1.8mm}{
\begin{tabular}{l|c|c|c|c}
\toprule
Dataset & \multicolumn{2}{c|}{SFRES} & \multicolumn{2}{c}{SFHOT} \\
\midrule
Metric & Nat. & Info. & Nat. & Info. \\
\midrule
\multicolumn{5}{c}{\textit{Trained Metric}} \\
\midrule
BARTScore (CNNDM) & 0.289 & 0.238 & 0.288 & 0.235 \\
UniEval (Summ) & 0.333 & 0.225 & \underline{0.320} & 0.249 \\
UniEval (Dial) & 0.291 & 0.194 & 0.291 & 0.196 \\
\midrule
\midrule
\multicolumn{5}{c}{\textit{Untrained Metric}} \\
\midrule
ROUGE-1 & 0.170 & 0.115 & 0.196 & 0.118 \\
ROUGE-L & 0.169 & 0.103 & 0.186 & 0.110 \\
MoverScore & 0.190 & 0.153 & 0.242 & 0.172 \\
BERTScore & 0.219 & 0.156 & 0.178 & 0.135 \\
BARTScore & 0.200 & 0.164 & 0.165 & 0.158 \\
CTRLEval & 0.195 & 0.177 & 0.121 & 0.158 \\
DecompEval (Ours) & \underline{\textbf{0.345}} & \underline{\textbf{0.242}} & \textbf{0.316} & \underline{\textbf{0.302}} \\
\bottomrule
\end{tabular}}
\caption{Spearman correlation of naturalness (Nat.) and informativeness (Info.) on data-to-text generation datasets. The highest correlation for each dimension achieved by untrained metrics is \textbf{bold}, while the highest correlation overall is \underline{underlined}.}
\label{tab:generalizationd2t}
\end{table}


\begin{table} [!t]
\centering
\scriptsize
\begin{tabular}{l|c|c}
\toprule
Dataset & Ours vs. UniEval (Summ) & Ours vs. UniEval (Dial) \\
\midrule
SummEval & \cellcolor{lime} 0.359 vs. 0.474  & \cellcolor{pink} 0.359 vs. 0.305 \\
Topical-Chat & \cellcolor{pink} 0.499 vs. 0.315 & \cellcolor{lime} 0.499 vs. 0.577 \\
SFRES & \cellcolor{pink} 0.293 vs. 0.279 & \cellcolor{pink} 0.293 vs. 0.243 \\
SFHOT & \cellcolor{pink} 0.309 vs. 0.285 & \cellcolor{pink} 0.309 vs. 0.244 \\
\bottomrule
\end{tabular}
\caption{Comparison of Spearman correlation averaged over all the dimensions in each dataset. Red indicates that our metric is better while green means the opposite.}
\label{tab:generalizationtask}
\end{table}

\subsection{Interpretability}
\label{sec:interpretability}

To verify whether the subquestions with their answers are reliable evidence to interpret the evaluation score, we conduct human evaluation on the generated answers to subquestions.
We randomly select 200 subquestions from each dimension of the Topical-Chat dataset.
Three annotators are hired to answer these subquestions with yes or no according to the evaluation input, where the human-annotated labels are determined via majority voting. The human-annotated labels are used as ground-truth labels to measure the quality of generated answers.

\begin{figure}[!t]
  \centering
  \includegraphics[width=1.0\linewidth]{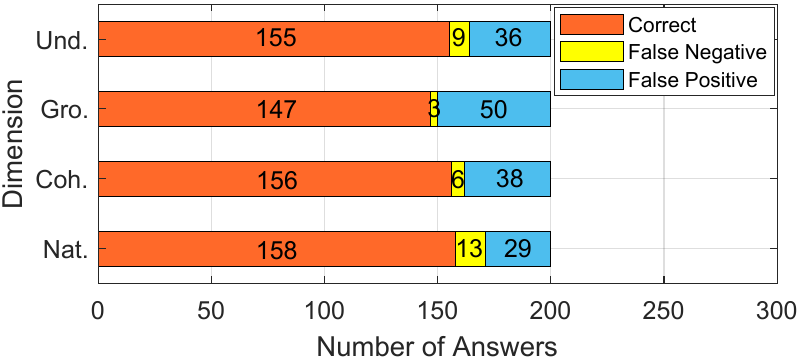}
  \caption{Human evaluation on subquestion answering in naturalness (Nat.), coherence (Coh.), groundedness (Gro.), and understandability (Und.) of Topical-Chat.}
  \label{fig:humaneval}
\end{figure}

\begin{figure}[!t]
  \centering
  \includegraphics[width=1.0\linewidth]{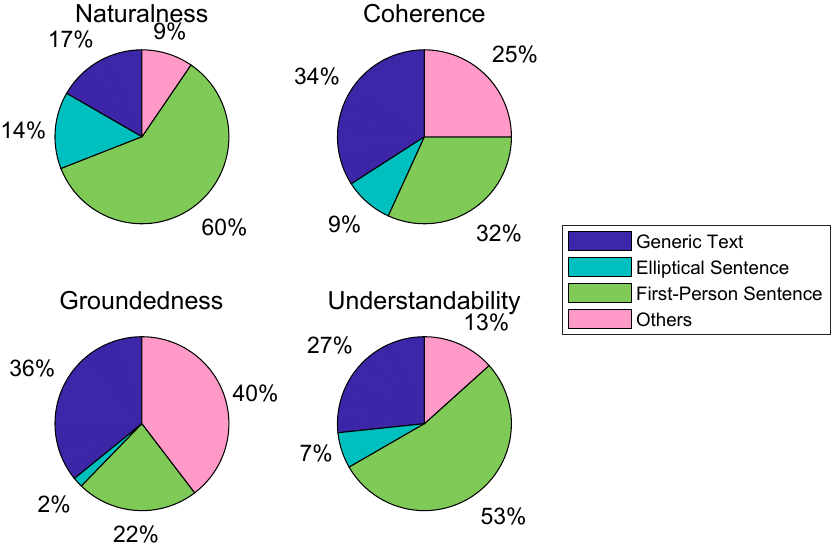}
  \caption{Type distributions of generated texts in the error cases of human evaluation.}
  \label{fig:errorcase}
\end{figure}

The results in Figure \ref{fig:humaneval} show that the accuracy of each dimension is above 0.7, indicating reasonable performance of subquestion answering which serves as interpretable evidence. We manually check the error cases and find that they include three typical types of generated texts, i.e., generic texts \cite{li2016diversity}, elliptical sentences, and first-person sentences. The type distributions of generated texts in the error cases are shown in Figure \ref{fig:errorcase}. We can observe that generic texts (such as "\textit{That is true.}") dominate the generated texts in the error cases of coherence / groundedness, while first-person sentences (such as "\textit{I think ...}") appear more frequently in those of naturalness / understandability. These two types of generated texts are mostly not contradictory to the evaluation input, thereby being commonly recognized by our metric. However, generic texts can only provide limited information while first-person sentences may contain irrelevant contents regarding the evaluation input. Thus, annotators tend to regard them as low-quality ones.

We also provide the case study in Appendix \ref{app:casestudy} to show the interpretable evaluation process of DecompEval.

\subsection{Ablation Study}
\label{sec:ablation}

\begin{table} [!t]
\centering
\small
\setlength{\tabcolsep}{1.8mm}{
\begin{tabular}{l|c|c|c|c}
\toprule
Metric & Nat. & Coh. & Eng. & Gro. \\
\midrule
DecompEval & \textbf{0.435} & \textbf{0.435} & \textbf{0.467} & \textbf{0.659} \\
\midrule
w/o Instruction & 0.427 & 0.418 & 0.442 & 0.651 \\
w/o Decomp. Q\&A & 0.411 & 0.399 & 0.433 & 0.643 \\
w/ Prefix Yes/No Que. & 0.431 & 0.402 & 0.461 & 0.601 \\
\bottomrule
\end{tabular}}
\caption{Spearman correlation of ablation models in naturalness (Nat.), coherence (Coh.), engagingness (Eng.), and groundedness (Gro.) of the Topical-Chat dataset.}
\label{tab:ablation}
\end{table}

To further investigate the effectiveness of each part in our metric, we conduct detailed ablation studies. We build the following three ablation models which remove three important parts of input prompts in \S\ref{sec:questionrecomp}, respectively: 1) \textit{w/o Instruction} indicates the model without the instruction $s$; 2) \textit{w/o Decomp. Q\&A} denotes the model without the decomposed subquestions with their answers $\{(sq_t,a_t)\}_{t=1}^n$; 3) \textit{w/ Prefix Yes/No Que.} means moving the yes/no question $q$ to the prefix of the evaluation input behind the instruction. We find that our metric without this yes/no question fails to achieve reasonable performance possibly because it contains the information about evaluation tasks and dimensions.


The results are shown in Table \ref{tab:ablation}. We can observe that all these three parts contribute to final performance. The decomposed subquestions with their answers play a more important role in most of the dimensions, indicating their positive impact as evidence on the model performance in addition to the interpretability. As for instructions, the performance of DecompEval without instructions does not degrade obviously. We conjecture that the yes/no question has explicitly conveyed the information to make the instruction-tuned PLM answer with yes or no. Thus, the impact of instructions may be weakened. 
The position of yes/no questions has also affected the model performance. From the experimental results, the question in the end of input prompts can obtain better performance than that in the middle part.


\subsection{Analysis on Model Scale}
\label{sec:modelscale}

\begin{table} [!t]
\centering
\small
\setlength{\tabcolsep}{1.4mm}{
\begin{tabular}{l|c|c|c|c|c}
\toprule
Base Model & \#Param & Nat. & Coh. & Eng. & Gro. \\
\midrule
FLAN-T5-Base & 250M & 0.175  & 0.206 & 0.386 & 0.291  \\ 
FLAN-T5-Large & 780M & 0.217 &  0.165 & 0.390  &  0.525 \\
FLAN-T5-XL & 3B & \textbf{0.435} & \textbf{0.435} & \textbf{0.467} & \textbf{0.659} \\
\bottomrule
\end{tabular}}
\caption{Spearman correlation of different base models in naturalness (Nat.), coherence (Coh.), engagingness (Eng.), and groundedness (Gro.) of Topical-Chat. \#Param means the number of model parameters.}
\label{tab:modelscale}
\end{table}

We further conduct experiments on the scale of base models, which may impact the capacity of following instructions to evaluate generated texts. We choose FLAN-T5-Base and FLAN-T5-Large additionally, and compare their performance with FLAN-T5-XL used in our main experiments.

The results in Table \ref{tab:modelscale} show that the performance of DecompEval improves on most of the dimensions as the number of parameters in the base model increases. We also find that there is a relatively large margin between the performance of FLAN-T5-Base/Large and FLAN-T5-XL, especially in the dimensions of naturalness, coherence, and groundedness. This phenomenon is accordant to the findings of existing works \cite{chung2022flant5,wei2022flan}, where the zero-shot capacity of instruction following mainly emerges in the models of sufficiently large scales.

\section{Discussion}

\noindent \textbf{Applicability in Non-English Languages}: Although the benchmark datasets in the experiment are mainly in English, our method can be also applied to non-English languages. Since our base model FLAN-T5 has some multilingual ability \cite{chung2022flant5}, we can design instruction-style questions / subquestions and answer words in the target language to apply DecompEval to non-English evaluation tasks. DecompEval can also adapt to stronger instruction-tuned multilingual PLMs for better applicability in non-English languages. We will further investigate the extensibility of our method to non-English evaluation tasks in the future work.

\section{Conclusion}

We present an untrained evaluation metric called DecompEval, which
formulates NLG evaluation as an instruction-style QA task, and utilizes instruction-tuned PLMs to solve this task via question decomposition.
Experimental results show that DecompEval achieves state-of-the-art performance in untrained metrics, which also exhibits better dimension-level / task-level generalization ability than trained metrics and improves the interpretability.

\section*{Limitations}

The limitation of our work includes the following aspects:

\noindent 1) The instruction-style question which measures the quality of generated texts from different dimensions still needs manual design. Although the questions in our experiment have already involved typical dimensions in text summarization, dialogue generation, and data-to-text generation, we admit that it is hard to cover all the dimensions in various NLG tasks. We believe that this is not a severe problem because we can refer to the definition and human annotation instructions \cite{mehri2020usr} of each dimension, which are commonly formulated as questions. We leave the exploration of automatically constructing instruction-style questions for multiple dimensions of NLG evaluation as future work.

\noindent 2) Due to the limitation of computational resources, the largest base model used in our experiment is FLAN-T5-XL with 3B parameters. Since the ability of instruction following is related to the model scale \cite{wei2022flan}, we leave the exploration of adopting larger instruction-tuned PLMs such as FLAN-T5-XXL and OPT-IML \cite{iyer2022optiml} as future work.



\section*{Acknowledgements}

This work was supported by the NSFC project (Key project with No. 61936010). This work was also supported by the Guoqiang Institute of Tsinghua University, with Grant No. 2020GQG0005.

\bibliography{custom}
\bibliographystyle{acl_natbib}

\appendix

\section{Prompt Design}
\label{app:instruct}

We show the specific prompt design for each dimension of SummEval, Topical-Chat, and SFRES / SFHOT in Table \ref{tab:promptsum}, \ref{tab:promptdialog}, and \ref{tab:promptd2t}, respectively. The instruction used in all the datasets is $s=$"\textit{Answer the following yes/no question.}", as mentioned in \S\ref{sec:mhqaformulation}. We refer to the definition and human annotation instructions of each dimension \cite{fabbri2021summeval,mehri2020usr,wen2015d2teval} as well as the existing works on QA for evaluation \cite{daniel2021qaeval,zhong2022unieval} to design evaluation inputs and yes/no questions. The format of subquestions is similar to yes/no questions, where the sentence to be measured is added to the middle part.

\begin{figure}[!t]
  \centering
  \includegraphics[width=1.0\linewidth]{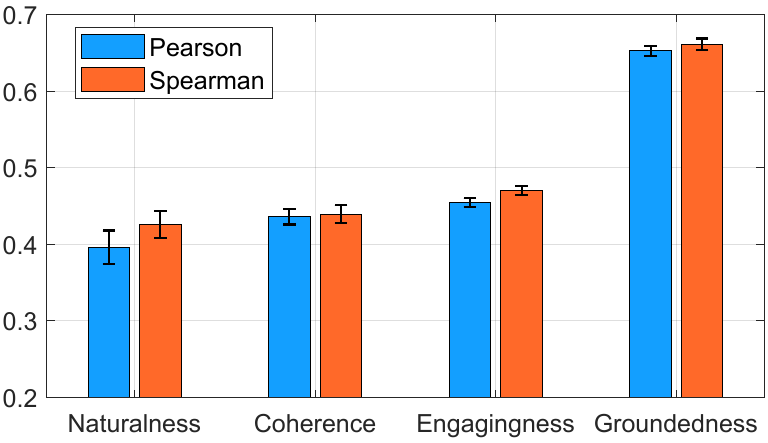}
  \caption{Mean values and standard deviations of Pearson and Spearman correlations over the original prompt and perturbed prompts on the Topical-Chat dataset.}
  \label{fig:variation}
\end{figure}

To investigate the sensitivity of input prompts, we construct seven grammatical yes/no questions for each dimension of Topical-Chat, covering the original one and three types of lexical variations, i.e., auxiliary verb replacement, synonym replacement, and word reordering. For example, the original yes/no question for naturalness in Table \ref{tab:promptdialog} is "\textit{Is this response natural to the dialogue history?}". After auxiliary verb replacement, the question may start with another auxiliary verb, such as "\textit{Does this response have a natural body to the dialogue history?}". Similarly, after synonym replacement, the question may have some words which are replaced with their synonyms, such as "\textit{Is this response natural given the dialogue history?}". As for word reordering, the question may be composed of reordered words, such as "\textit{Is this a natural response to the dialogue history?}". Note that the subquestions are perturbed accordingly. Then, we illustrate the mean value and standard deviation over the original prompt and perturbed prompts of each dimension in Figure \ref{fig:variation}, showing the stable performance of DecompEval faced with variations.

\begin{table*} [!h]
\centering
\small
\begin{tabular}{l|l|l|l}
\toprule
Dimension  & Evaluation Input & Yes/No Question & Subquestion \\
\midrule
Coherence  & document: $c$ & Is this a coherent summary to the  & Is this summary sentence $t$ $x_t$ a coherent \\
& summary: $x$ & document? &  summary to the document? \\
\midrule
Consistency & claim: $x$ & Is this claim consistent with the  & Is this claim sentence $t$ $x_t$ consistent with \\
& document: $c$ & document? & the document? \\
\midrule
Fluency & paragraph: $x$ & Is this a fluent paragraph? &  Is this paragraph sentence $t$ $x_t$ a fluent  \\
& & & paragraph? \\
\midrule
Relevance & summary: $x$ & Is this summary relevant to the  &  Is this summary sentence $t$ $x_t$ relevant to  \\
& reference: $r$ & reference? & the reference? \\
\bottomrule
\end{tabular}
\caption{Input prompt design for each dimension of the SummEval dataset, including the evaluation inputs ($c,x,r$), yes/no questions ($q$), and decomposed subquestions ($\{sq_t\}_{t=1}^n$).}
\label{tab:promptsum}
\end{table*}

\begin{table*} [!h]
\centering
\small
\begin{tabular}{l|l|l|l}
\toprule
Dimension  & Evaluation Input & Yes/No Question & Subquestion \\
\midrule
Naturalness &   dialogue history: $c_{his}$  &  Is this response natural to the dialogue  & Is this response sentence $t$ $x_t$   \\
& response: $x$ &  history? & natural to the dialogue history? \\
\midrule
Coherence  &   dialogue history: $c_{his}$   &  Is this a coherent response given the  &  Is this response sentence $t$ $x_t$ a  \\
&  response: $x$  &  dialogue history? & coherent response given the \\
& & & dialogue history? \\
\midrule
Engagingness   &  dialogue history: $c_{his}$  & Is this an engaging response according  & Is this response sentence $t$ $x_t$ an\\
& fact: $c_{fact}$  & to the dialogue history and fact? &  engaging response according to \\
& response: $x$  & & the dialogue history and fact? \\
\midrule
Groundedness  &   response: $x$  & Is this response consistent with   &  Is this response sentence $t$ $x_t$ \\
& fact: $c_{fact}$   & knowledge in the fact?  & consistent with knowledge in the \\
& &  &  fact? \\
\midrule
Understandability   & dialogue history: $c_{his}$   & Is this an understandable response  & Is this response sentence $t$ $x_t$ an \\
 & response: $x$ &  given the dialogue history? & understandable response given the \\
 & & &  dialogue history? \\
\bottomrule
\end{tabular}
\caption{Input prompt design for each dimension of the Topical-Chat dataset, including the evaluation inputs ($c,x,r$), yes/no questions ($q$), and decomposed subquestions ($\{sq_t\}_{t=1}^n$). Note that Topical-Chat is a knowledge-grounded dialogue generation dataset, where the context $c$ contains dialogue histories $c_{his}$ and knowledge facts $c_{fact}$.}
\label{tab:promptdialog}
\end{table*}

\begin{table*} [!h]
\centering
\small
\begin{tabular}{l|l|l|l}
\toprule
Dimension  & Evaluation Input & Yes/No Question & Subquestion \\
\midrule
Naturalness & utterance: $x$ & Is this a fluent utterance? & Is this utterance sentence $t$ $x_t$ a fluent utterance? \\
\midrule
Informativeness & sentence: $x$ & Is this sentence informative & Is this sentence $t$ $x_t$ informative according to the  \\
& reference: $r$ & according to the reference? &  reference? \\
\bottomrule
\end{tabular}
\caption{Input prompt design for each dimension of the SFRES / SFHOT dataset, including the evaluation inputs ($c,x,r$), yes/no questions ($q$), and decomposed subquestions ($\{sq_t\}_{t=1}^n$).}
\label{tab:promptd2t}
\end{table*}

\begin{table*} [!t]
\centering
\small
\begin{tabular}{l|l}
\toprule
  Dialogue History & ... \\
  & Speaker A: I don't watch them very often. Apparently there was a showing of the recent film in a  \\
  & park in D.C. That's one U.S. city I haven't been to. \\
  & Speaker B: Sadly, I haven't been to DC either, although I've always wanted to visit there. \\
  & Apparently there's a lot of interesting going down this summer. They're having a crab feast at the  \\
  & Navy-Marine Corps Stadium. They'll have 100 gallons of crab soup! Can you imagine that much \\
  & soup? \\
\midrule
  Generated Response &  Wow that's a lot of soup. Are you talking about the Fort-Reno Concert? I heard flasher will \\
  &  perform there. \\
  \midrule
  Evaluation Dimension & Coherence \\
  \midrule
  Human Score (1-3) &  2.667 \\
  \midrule
  MoverScore (0-1) & 0.506  \\
  BARTScore (<0) & -3.867  \\
  CTRLEval (<0)  &  -4.768   \\
  UniEval (Dial) (0-1)  & 0.999  \\
  \midrule
  DecompEval (0-1)  &  0.855  \\
  \quad w/ Evidence & Is this response sentence 1 "Wow that's a lot of soup." a coherent response given the dialogue \\
  &  history? \quad \textbf{Yes} \\
  & Is this response sentence 2 "Are you talking about the Fort-Reno Concert?" a coherent response  \\
  & given the dialogue history? \quad \textbf{No} \\
  & Is this response sentence 3 "I heard flasher will perform there." a coherent response given the \\
  & dialogue history? \quad \textbf{Yes} \\
\bottomrule
\end{tabular}
\caption{Case study on the evaluation of coherence in the Topical-Chat dataset. The content in the bracket indicates the scale of evaluation scores in each metric, where higher scores mean better quality. The evidence of DecompEval denotes the subquestions with their answers.}
\label{tab:casestudy_dial}
\end{table*}

\begin{table*} [!t]
\centering
\small
\begin{tabular}{l|l}
\toprule
  Document & A southern Iowa chiropractor accused of accepting sex as payment for his services and  \\
  & performing exorcisms on patients has surrendered his state license ... \\
\midrule
  Generated Summary & A chiropractor in iowa has surrendered his license to practice and admitted to swapping services \\
  &  for sex and performing exorcisms on some patients. Manuel also recommended that patients   \\
  & stop taking medication no longer exist before he can resume practicing chiropractic in the state.   \\
  & The disgraced chiropractor received a perfect five out of five stars in patient satisfaction. \\
  \midrule
  Reference Summary & Charles Manuel of Lamoni, Iowa admitted to a review board that he traded sexual favors for his \\
  & services. Manuel also fessed up to performing exorcisms and to telling patients to stop taking  \\
  & medications prescribed to them by a medical doctor. The Iowa Board of Chiropractic required  \\
  & Manuel to pledge he would not apply for reinstatement of the license, but only for 10 years.\\
  \midrule
  Evaluation Dimension & Relevance \\
  \midrule
  Human Score (1-5) &  3.667 \\
  \midrule
  MoverScore (0-1) & 0.546  \\
  BARTScore (<0) &  -5.188 \\
  CTRLEval (<0)  &  -2.912   \\
  UniEval (Summ) (0-1)  & 0.060  \\
  \midrule
  DecompEval (0-1)  &  0.586  \\
  \quad w/ Evidence & Is this summary sentence 1 "A chiropractor in iowa has surrendered his license to practice and\\
  &  admitted to swapping services for sex and performing exorcisms on some patients." relevant to \\
  &  the reference? \quad \textbf{Yes} \\
  & Is this summary sentence 2 "Manuel also recommended that patients stop taking medication no    \\
  & longer exist before he can resume practicing chiropractic in the state." relevant to the reference? \\
  &  \textbf{Yes} \\
  & Is this summary sentence 3 "The disgraced chiropractor received a perfect five out of five stars in  \\
  & patient satisfaction." relevant to the reference? \quad \textbf{No} \\
\bottomrule
\end{tabular}
\caption{Case study on the evaluation of relevance in the SummEval dataset. The content in the bracket indicates the scale of evaluation scores in each metric, where higher scores mean better quality. The evidence of DecompEval denotes the subquestions with their answers.}
\label{tab:casestudy_sum}
\end{table*}

\section{Case Study}
\label{app:casestudy}

We provide evaluation cases on the Topical-Chat and SummEval datasets in Table \ref{tab:casestudy_dial} and \ref{tab:casestudy_sum}, respectively. We can observe that DecompEval can provide the evaluation scores which are the most accordant to human scores. Also, the subquestions with their answers can act as evidence to indicate the potential low-quality sentence which impacts the overall quality. For example, in Table \ref{tab:casestudy_dial}, the second sentence which mentions the concert seems not to be coherent to the topic in the dialogue history (i.e., the crab feast at a stadium). Similarly, in Table \ref{tab:casestudy_sum}, the third sentence about patient satisfaction is not relevant to the reference. In comparison, the evaluation scores of other metrics deviate from human scores, while they cannot provide evidence to demonstrate how they predict the evaluation scores.

\section{Analysis on Decomposition Strategy}

To judge how useful our decomposed subquestions with generated answers are for interpreting final evaluation scores, we ask the same three annotators to assign an interpretability score to selected samples in \S\ref{sec:interpretability}. We adopt a 1-3 Likert scale, where 1 / 2 / 3 means that the decomposition can hardly / partly / comprehensively help understand how the model reaches final scores, respectively. The average interpretability scores over all the selected samples are 2.84 / 2.76 / 2.74 / 2.67 for naturalness / coherence / groundedness / understandability, respectively, showing that our decomposition strategy based on sentences is mostly useful for interpreting final evaluation scores of multiple dimensions.

\section{Experimental Detail}

\subsection{License of Datasets and Models}

The licenses of datasets and base models used in our experiments include MIT for the SummEval dataset and Apache-2.0 for the Topical-Chat dataset and the FLAN-T5 model.

\subsection{Implementation Detail}

We use NLTK\footnote{https://www.nltk.org} to split generated texts into sentences for the construction of subquestions. As for the computation of Pearson, Spearman, and Kendall correlation coefficients, we use the APIs from SciPy\footnote{https://scipy.org}.

\subsection{Inference Time}

The inference time on the SummEval / Topical-Chat dataset is about 28 / 5 minutes, respectively. We test our model on 1 NVIDIA A100 GPU.

\subsection{Human Evaluation}

The annotation instruction of human evaluation in \S\ref{sec:interpretability} contains two main parts: 1) A subquestion with its corresponding instruction and evaluation input in the same format as Figure \ref{fig:overview}; 2) An explanation of NLG tasks and dimensions to be measured, which is re-printed from the original paper about benchmark datasets \cite{mehri2020usr}. In addition, all the other contents shown to annotators are from the original dataset of Topical-Chat \cite{gop2019topicalchat}. We manually check these contents before annotation to avoid potential risks.

We recruit three graduate students as annotators to complete this task. We pay each annotator \$0.07 for every subquestion. The payment is determined based on the difficulty of tasks and the length of subquestions.


\end{document}